\RequirePackage{silence}
\documentclass[runningheads]{llncs}

\usepackage{xcolor}
\usepackage{graphicx}
\usepackage{cite}
\usepackage{tikz}
\usepackage{array}
\usepackage{textcomp}
\usepackage{xcolor}
\usepackage{multirow}
\usepackage{booktabs}
\usepackage{float}
\usepackage{algorithmic}
\usepackage{lmodern}
\usepackage{pifont}
\usepackage{mathtools}
\usepackage{breqn}
\usepackage{lmodern}
\usepackage{multirow,tabularx}
\usepackage{flushend}
\usepackage{gensymb}
\usepackage{xurl}

\usepackage[vlined, ruled, shortend]{algorithm2e}
\graphicspath{{/}}
\newcolumntype{B}{>{\hsize=1.2\hsize}X}
\newcolumntype{S}{>{\hsize=.5\hsize}X}
\newcolumntype{T}{>{\hsize=.3\hsize}X}
\newcolumntype{Y}{>{\centering\arraybackslash}X}
\newcolumntype{y}{>{\centering\arraybackslash}S}
\newcolumntype{t}{>{\centering\arraybackslash}T}
\newcolumntype{R}{>{\raggedright}S}

\begin{document}
\title{Vision-based Safe Autonomous UAV Docking with Panoramic Sensors}
\author{
	Phuoc Nguyen Thuan\inst{1,2} \and
	Tomi Westerlund\inst{1} \and
	Jorge Peña Queralta\inst{1}
}%
\authorrunning{P. Nguyen Thuan et al.}
\titlerunning{Vision-based Safe Autonomous UAV Docking}
\institute{
	Turku Intelligent Embedded and Robotic Systems \\
	University of Turku, Finland\\
	\and
	Computing Sciences, Faculty of Information Technology and Communication Sciences, Tampere University, Finland \\[+.42em]
	\email{\{tpnguy, tovewe, jopequ\}@utu.fi}
}
\maketitle%
%
%
\begin{abstract}

	The remarkable growth of unmanned aerial vehicles (UAVs) has also sparked concerns about safety measures during their missions. To advance towards safer autonomous aerial robots, this work presents a vision-based solution to ensuring safe autonomous UAV landings with minimal infrastructure. During docking maneuvers, UAVs pose a hazard to people in the vicinity. In this paper, we propose the use of a single omnidirectional panoramic camera pointing upwards from a landing pad to detect and estimate the position of people around the landing area. The images are processed in real-time in an embedded computer, which communicates with the onboard computer of approaching UAVs to transition between landing, hovering or emergency landing states. While landing, the ground camera also aids in finding an optimal position, which can be required in case of low-battery or when hovering is no longer possible. We use a YOLOv7-based object detection model and a XGBooxt model for localizing nearby people, and the open-source ROS and PX4 frameworks for communication, interfacing, and control of the UAV. We present both simulation and real-world indoor experimental results to show the efficiency of our methods.


\end{abstract}

\section{Introduction}

Recently, unmanned aerial vehicles (UAVs, or drones) have seen an unprecedented rise in their adoption rate, primarily thanks to technological advancements improving their availability and dependability~\cite{nex2022uav}. They have been vital components in multiple civil applications, ranging from remote sensing applications~\cite{xiang2011uavremotesensing} to aerial delivery~\cite{song2018dronedeli}.

One of the key issues stopping wider adoption of UAVs for civilian applications in urban areas is safety and security~\cite{milano2022air}. Autonomous UAVs flying over populated areas pose inherent hazards. The risk increases significantly during take-off and docking maneuvers, with potential risks for nearby passers. This paper seeks to address the safety of persons near a landing area and define a framework for safety-aware autonomous UAV landing with minimal ground infrastructure.

Specifically, the aim is to first design and develop a solution with minimal infrastructure footprint and commercial off-the-self components. Then, we validate the functionality of the system through a series of experiments in the Gazebo simulator and our $9\times 8\times 5$\,m indoor test area. Our goal is to provide a solution that can further enhance the safety of autonomous UAV landing operations. One of the fundamental aspects of UAV landing safety is the avoidance of potential hazards on the landing path. While the concept of hazards avoidance during UAV landing is vast, we narrow the scope to protecting pedestrians near the landing area.

\begin{figure}[!htb]
	\centering
	\includegraphics[width=\textwidth]{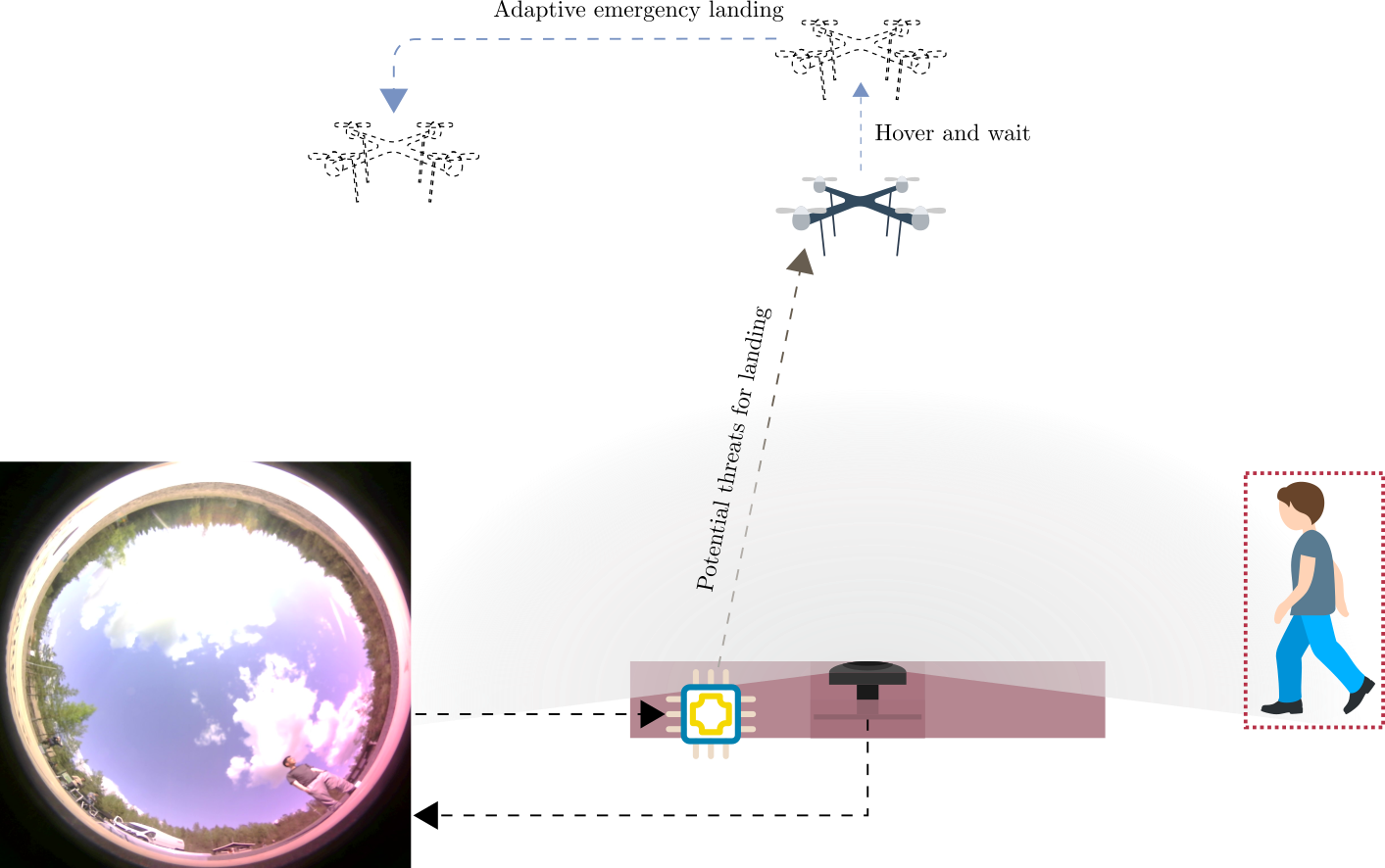}
	\caption{Envisioned system.}
	\label{fig:envision}
\end{figure}

Leveraging on the recent rapid development of deep-learning-enabled computer vision on embedded hardware~\cite{bhowmik2018embeddedvision} and the high potential of 360\degree\ panoramic sensors, we approach the problem with an on-ground vision-based system for landing area monitoring to identify people who are at risk from the landing UAV. Our envisioned system is a lightweight landing pad with a single panoramic camera in the center providing a bottom-view that gives the system a 360\degree\ view of the surroundings. An embedded computing unit processes the information to generate relevant information, and packages them as lightweight, efficient messages to send to the UAV to adjust its landing trajectory. Figure~\ref{fig:envision} illustrates our envisioned system and the intended behavior.

On-ground approach for safe UAV landing have two significant advantages over its onboard counterpart. First, it widens the options for computing platforms and sensors. UAV payloads are limited, so for tasks such as aerial delivery, every gram of weight that can be saved by replacing heavy companion computers and sensors with more lightweight options is directly transferred to the weight that their primary task requires them to carry. The solution described and implemented in this paper does not involve a very high-end computing platform.  Second, ground-based solutions are potentially more robust to limited environment observability from UAVs, and can also serve as a redundant way of ensuring safety in such critical scenarios.

Moreover, we design and implement the safe UAV landing software based on open-source libraries. Our software components include the detection module, which consists of an object detector and a distance estimator to identify and localize people in a two-dimensional space, and an autonomous flight program that safely allows the UAV to land while maintaining complete autonomy by using the information about the surroundings provided by the detection module. The functionality of each software component and the communication between them is facilitated by the free and open-source Robot Operating System(ROS), which has become the de facto standard for robotic applications in recent years. The popular autopilot library PX4 is also utilized for high-level UAV control and integration of autonomous flight algorithms.

The rest of this paper is organized as follows. Section 2 introduces related works in computer vision for panoramic sensors, and vision-based approaches used in UAV landing. Section 3 introduces our methodology for a ground-based vision-based safe UAV landing framework. Section 4 then reports our experimental setup and results. Finally, Section 5 concludes the work.

\section{Related Works}
\label{sec:relworks}
\subsection{Vision-based systems for autonomous UAV landing}
In the literature, research for vision-enabled autonomous landing systems for UAVs, primarily multi-rotor vehicles, can be divided into two main categories: onboard and on-ground. According to the survey by~\cite{kong_vision-based_survey_2014}, the former approach is the more predominant and well-studied approach, with multiple systems developed for landing on known, unknown and moving areas, while works done for on-ground vision-based landing systems are still scarce.

A common point among on-ground systems is that they utilize a diverse range of sensing units because these systems are not restricted by UAV payload. However, most of the work in this category focuses on the pose estimation and control of the UAV rather than the monitoring of the landing site. In one of the earlier research on on-ground monitoring systems,~\cite{wang2006autonomous} introduced a computer control camera platform to identify square markers with known size patched on micro aircraft to measure their three-dimensional coordinates. The main limitation of this method was the camera's narrow field of view (FOV) and reliance on a step motor to shift its orientation to access other viewpoints. ~\cite{martinez2009trinocular} later introduced a system that can estimate UAV's position based on onboard key features in real-time by extracting information provided by a trinocular camera system on the ground. Alternatively, instead of standard RGB cameras, ~\cite{yang_ground-based_2016} presented a ground-based guidance system utilizing an array of near-infrared cameras, which significantly increases the detection range to detect, track, and autonomously land a fixed-wing UAV without reliance on GPS data.

Several other works have also presented onboard methods that select safe landing zones by detecting potential hazards on the landing path~\cite{alam2021surveyslz}. In these papers, the authors utilize lightweight convolutional neural networks such as YOLO~\cite{safadinho2020uavcvlanding} and MobileNet~\cite{castellano2020crowd} to detect safe landing zones, which are away from individual or groups of people in populated areas~\cite{tovanche2022visual}, or flat and obstacle-free areas~\cite{marcu2018safeuav}.

\subsection{Object detection on panoramic images}
\label{subsec:fishobjdet}
Object detection on panoramic images is a topic that is also less well-studied than its pinhole counterpart within the literature. One concept that has been researched to adapt object detection models to fisheye imagery, which frequently has oriented and radially distorted objects, is alternative representations for standard bounding boxes. ~\cite{rashed2021fisheyeyolo} explored the usage of curved boxes, oriented boxes, ellipses, and polygons. YOLOv3~\cite{redmon2018yolov3} was adapted and modified to output these different representations. The results show that 24-sided polygons achieved the most reasonable tradeoffs between model complexity and accuracy. Further analysis also reports no drops in inference speed when increasing the number of vertices. Alternatively, ~\cite{xu2021gliding} proposed a simple framework for oriented box representation by gliding each vertex of the original horizontal box on its corresponding side to get more accurate coverage of the detected object and demonstrated the method's effectiveness in object detection on aerial images, texts, and pedestrians in fisheye images.

~\cite{zhu2019object} presented a localization method by leveraging top-view fisheye images from a UAV and altitude data. The proposed framework first involves acquiring pixel positions of objects using an object detection model implemented based on the RetinaNet model~\cite{lin2017focalretinanet} with MobileNet~\cite{mobilenet} backbone for more efficient computing. Then, by fusing the camera's parameter and height data from other sensors, a series of coordinate transforms is performed to obtain the object's position in world coordinates.

In addition, some public fisheye image datasets were published to facilitate the development of this field of research. Most noticeable is the Woodscape dataset~\cite{woodscape} for autonomous driving, comprising over 100,000 images from four surrounding cameras. Later, in 2022, the KITTI-360 dataset~\cite{liao2022kitti360} was released as a successor to the popular KITTI dataset~\cite{kitti}. It expanded on the original work with more data for suburban driving from multiple sensor units, including two 180\degree\ fisheye cameras on each side of the station wagon.

\section{Methodology}
\subsection{Detection module}
\label{sec:detmod}
To identify whether the landing spot is safe, we implement a system that detects people around the area and estimates their distances to the camera. For the rest of this work, we will refer to this combination of human detection and distance estimation system as the \textit{detection module}.

\subsubsection{Human detection}
\label{subsec:yoloimp}
This project's object detection model is based on YOLOv7 \cite{yolov7-wang2022}, whose official project provides different model versions with varying sizes and complexity. The standard models are the tiny version, which optimizes for high throughput and minimal footprint to run on edge GPU; the normal version, namely YOLOv7, for regular consumer-grade GPUs; and the more powerful, cloud GPU-oriented YOLOv7-W6. To further optimize YOLOv7-tiny for edge GPUs, the authors use Rectified Linear Unit (ReLU) as the activation function. On the other hand, for other versions, Sigmoid-weighted Linear Unit (SiLU) \cite{elfwing2018silu} is used as the activation function. For this work, we mainly consider the tiny and the normal versions of YOLOv7, as empirical testing shows they are more suitable for deployment on our embedded platform.

\subsubsection{Distance estimation}
\label{subsec:distest}
Monocular depth estimation is a challenging topic that has received much attention recently. The most common approach is to train a deep learning model to predict depth from an arbitrary input image \cite{zhao2020monocular, ranftl2020towards}. The training data can be from multiple measuring tools like LIDAR, RGB-D, and stereo cameras. Unfortunately, most public datasets only have depth images in perspective view. To our knowledge, no available pre-trained monocular depth estimation models trained on data with the same characteristics as ours exist. Another possible approach that has been studied is integrating a distance estimator head into the object detector's architecture \cite{vajgl2022distyolo}.

\begin{figure}[!ht]
	\centering
  \includegraphics[width=0.7\textwidth]{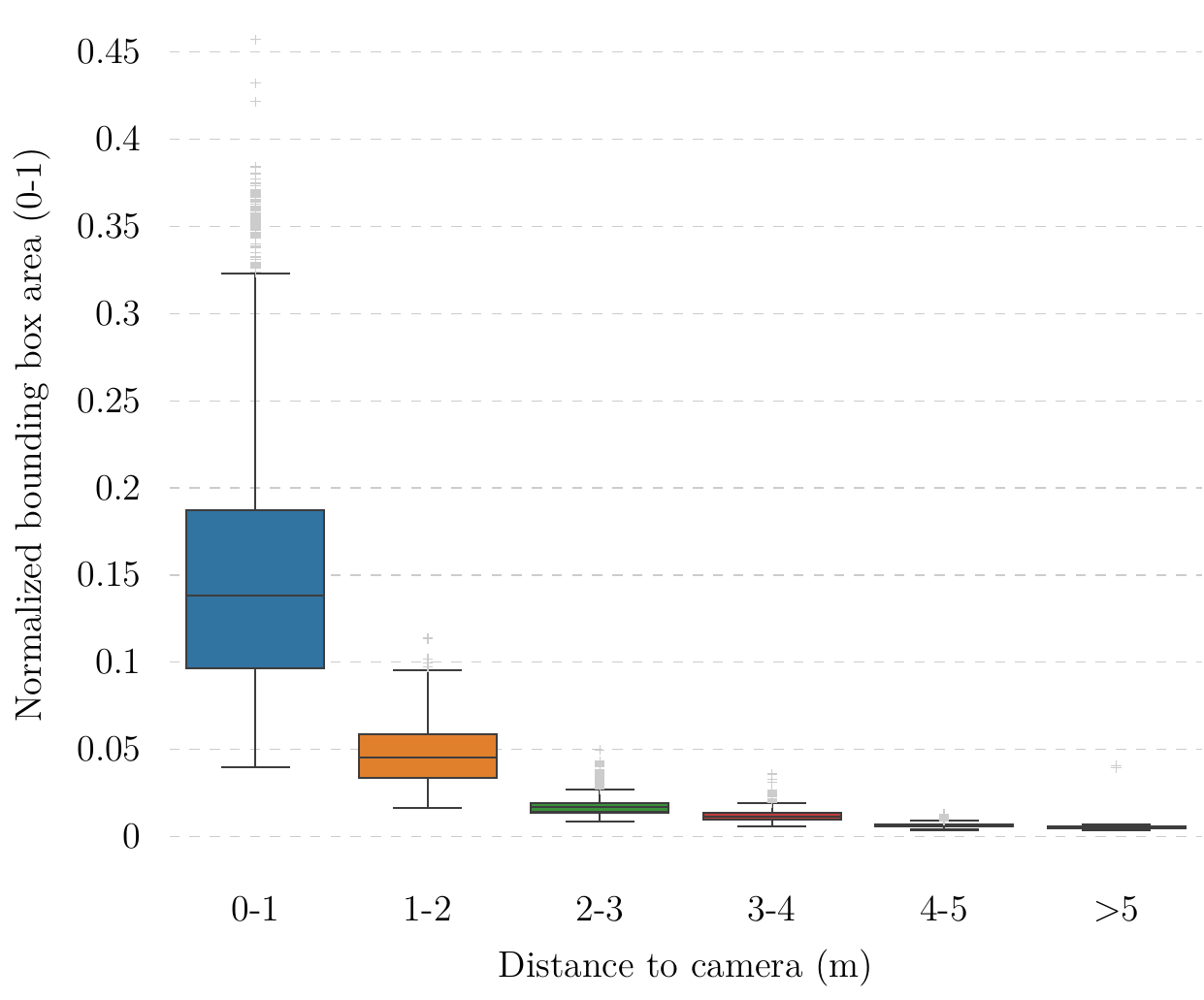}
	\caption{Relationship between normalized bounding box area and distance between the object to the camera. It is worth noting that high position accuracy is not needed; instead, a high recall in the detection (low probability of false negatives) and good classification accuracy (safe or unsafe distances) are more important.}
	\label{fig:distarearel}
\end{figure}

The goal for the system is not to prioritize precisely predicting the distance of the person to the camera but instead to get a rough estimate of whether the person is close or far away from the camera to determine if the surrounding area is safe for UAV landing. Therefore, we choose a more straightforward solution that integrates well with the rest of our system and requires little computational power during inference time. Specifically, we leverage the bounding boxes information from the object detector as input for a regression model to predict the people's distance to the camera. The regression model of choice is the gradient-boosted decision trees algorithm implemented with the XGBoost library \cite{chen2016xgboost}. While previously shown in Figure~\ref{fig:distarearel} that the bounding box areas are correlated to the distance, better results can be obtained when inferring with other bounding box details, including its center point coordinates $(x,y)$ and its dimensions $(w,h)$, since two pictures showing the same person at the same distance to the camera can have much different bounding box shapes when the orientation changes. Furthermore, varying poses, e.g., crouching and sitting, can drastically change the shapes of the bounding boxes as well.

\subsubsection{Vision-based localization}
\label{subsec:vislocal}
\begin{figure}[!htb]
	\centering
	\includegraphics[width=0.6\textwidth]{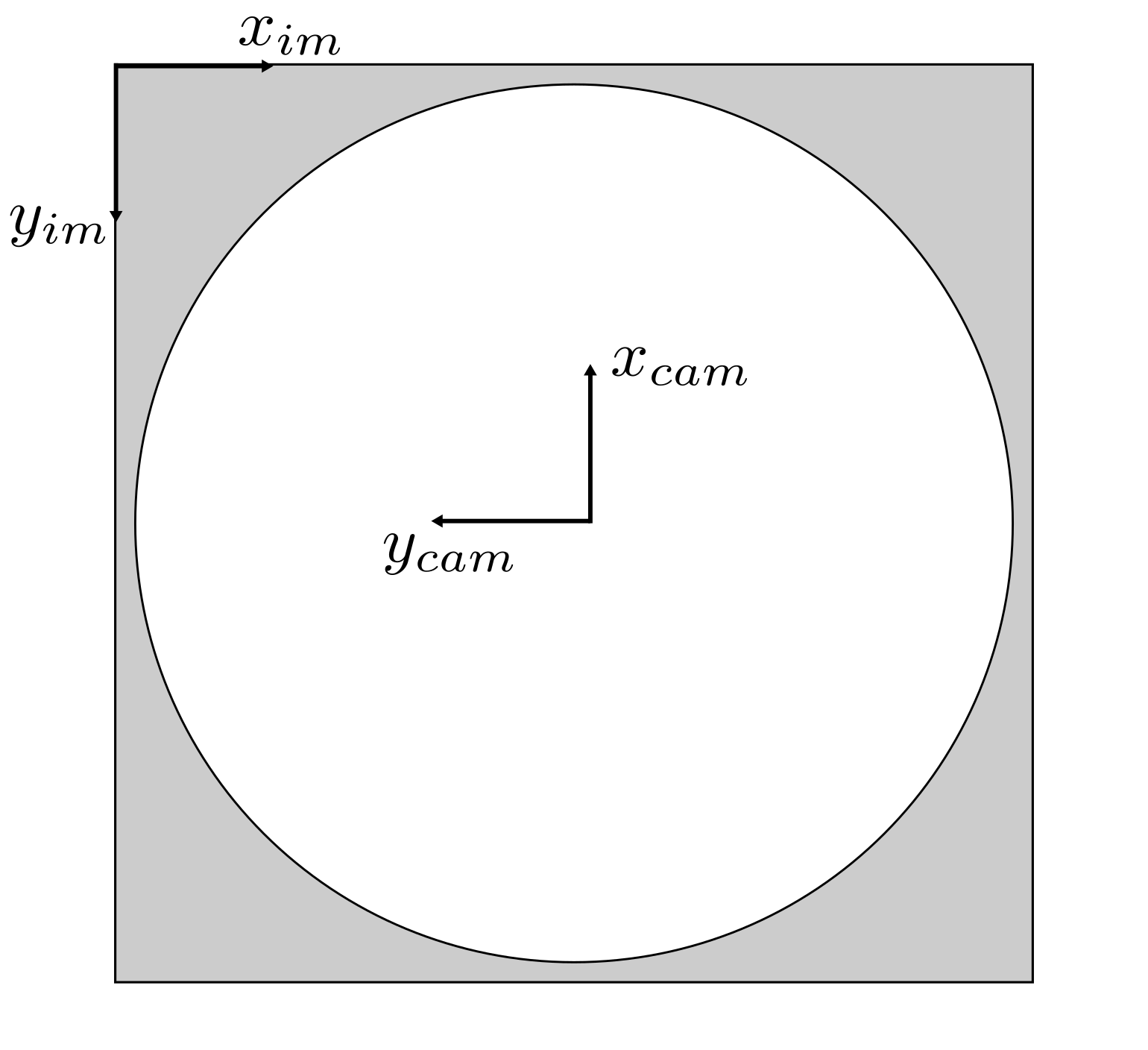}
	\caption{Image and camera coordinate systems.}
	\label{fig:coordchange}
\end{figure}

The bounding boxes from the object detector can provide insight into the relative orientation of a detected person to the camera, and the distance predicted by the XGBoost model can estimate how far they are from the camera. Fusing these two pieces of information allows a person to be sufficiently localized in a two-dimensional space. Initially, the image coordinate system must be transformed to one that matches the camera's coordinate system. To simplify the experiment, we position the camera and vehicle to align their coordinate axes with the world coordinate system (in this case, the coordinates of the MOCAP system). In the standard image coordinate system, the origin lies in the top left corner, with a horizontal x-axis from left to right and a vertical y-axis pointing downwards. To transform the image coordinates (in pixels) to the camera coordinate system depicted in Figure~\ref{fig:coordchange}, the transformations are as follows:
\begin{align}
	\begin{split}
		x_{cam\_pix} = -y_{image} + \frac{w_{image}}{2} \\
		y_{cam\_pix} = -x_{image} + \frac{h_{image}}{2}
	\end{split}
\end{align}

Suppose the camera is not aligned with the world coordinate system. In that case, the offset angle between the world's and the camera's coordinate system must be pre-known, and a two-dimensional rotation transformation must be performed to obtain the coordinates of the detected people with respect to the world coordinate system on which the vehicle's local position and orientation are based.

\begin{figure}[!htb]
	\centering
	\includegraphics[width=\textwidth]{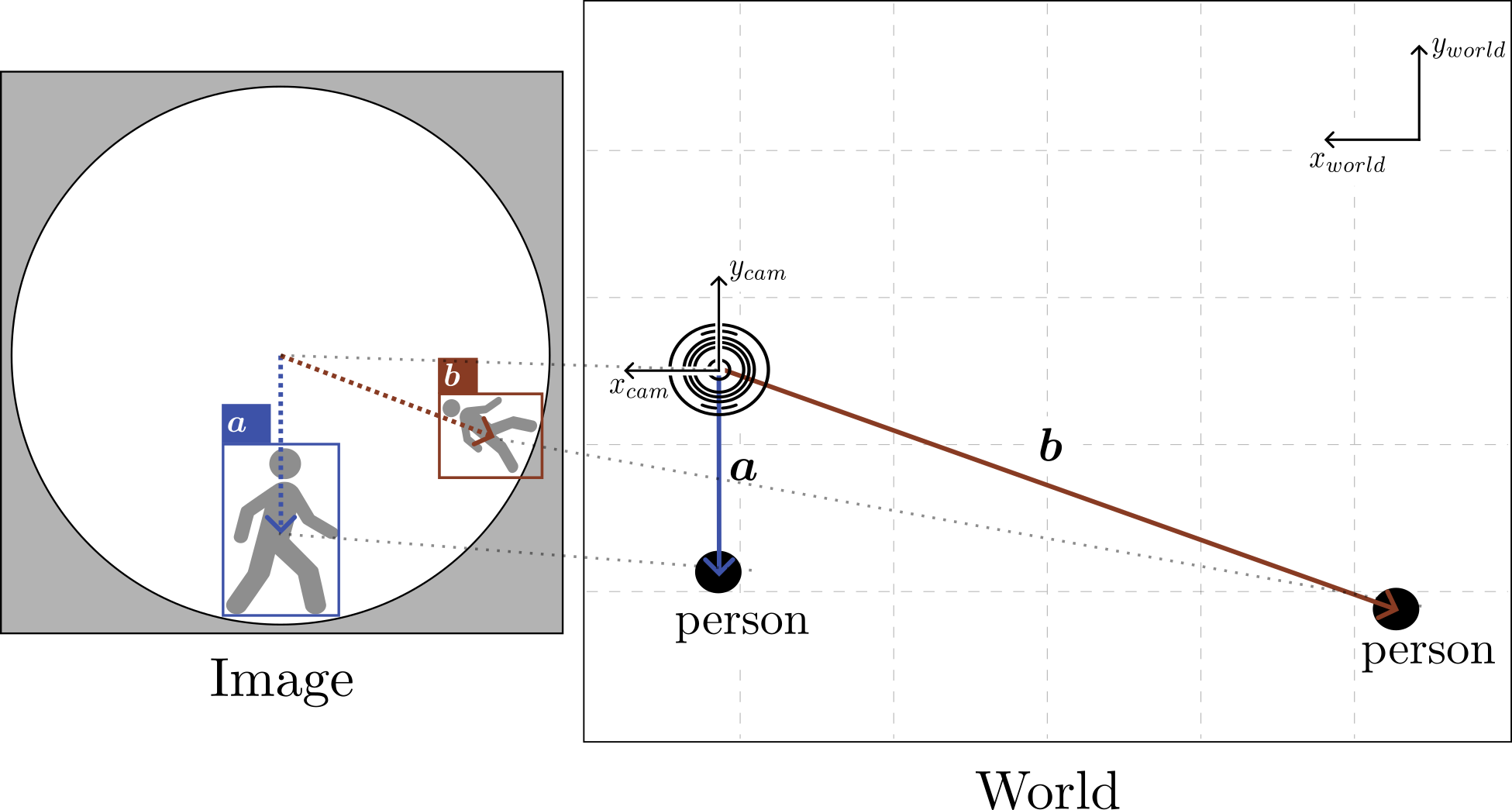}
	\caption{Illustration of how image coordinates are transformed to positions of detected people. Because of the panoramic nature of the sensor, images are heavily distorted near the ground level or image edges.}
	\label{fig:im2world}
\end{figure}

From the transformed coordinate system, the orientation of the directional vector pointing to detected people can be obtained. For simplicity, we define this directional vector as the normalized vector from the camera coordinates origin to the center point of the corresponding bounding box. Then, the predicted distance from the XGBoost model is used to scale this vector to an approximate position of the detected person in two-dimensional space. We denote the coordinates of a person in the world coordinate system as $X_p$ for simplicity, and the formula for calculating it is:

\begin{align}
	X_p=\begin{bmatrix}x_{p} \\ y_{p}\end{bmatrix} = d_{pred} \frac{\begin{bmatrix}x_{cam\_pix} \\
			y_{cam\_pix}\end{bmatrix}}{\left\lVert \begin{bmatrix}x_{cam\_pix} \\ y_{cam\_pix}\end{bmatrix}
		\right\rVert} +
	\begin{bmatrix}d_{x\_cam} \\ d_{y\_cam}\end{bmatrix}
\end{align}
$d_{x\_cam}$ and $d_{y\_cam}$ are the camera's position in the world coordinate system, and $d_{pred}$ is the distance prediction result from the distance estimator. Figure~\ref{fig:im2world} summarizes the localization process, from acquiring the people's coordinates in a frame to projecting them into the world coordinates.

\subsection{Safe landing program}
\label{sec:autonomousflight}
\subsubsection{System behavior}
\begin{figure}[!htb]
	\centering
	\includegraphics[width=\textwidth]{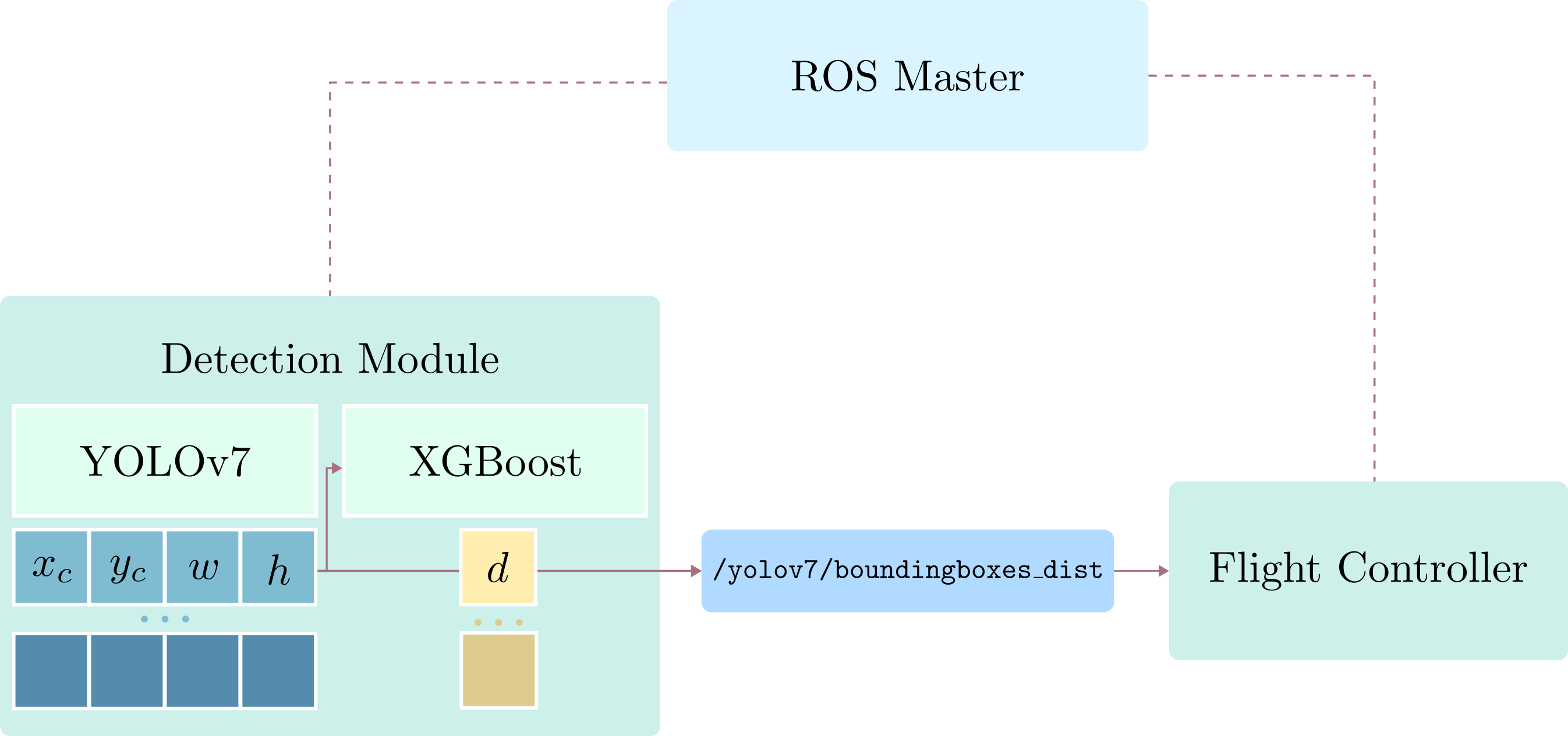}
	\caption{Communication between the detection module and the flight controller over a ROS network. Only the detection and position estimation results are sent to the UAV.}
	\label{fig:detnode2uav}
\end{figure}

In our deployment, both in the Gazebo simulation and in a real-world indoor experiment, the vehicle operates in \textit{offboard} mode, which allows it full autonomy, with \textit{position} mode as the fallback in case of failure. In indoor environments, the UAV utilizes local coordinates for localization and determining mission setpoints. The experimental flight mission consists of four phases: taking off, performing the flight mission, pre-landing, and landing. The first, second, and last phases are self-explanatory, while the pre-landing phase activates the safe landing mechanism. We define pre-landing as going to a setpoint at a height safe for complete landing while continuously communicating over a ROS network with the detection module on the ground for information of the surrounding as shown in Figure~\ref{fig:detnode2uav}.

During pre-landing, the vehicle will retreat to a safe position and hover if the detection module detects a person within a predefined safe threshold. After a set period, the vehicle switches to adaptive emergency landing mode and searches for an optimal landing position. The vehicle can also resort to this behavior in circumstances where hovering is impossible, e.g., when the payload is over a threshold or when the battery is low. We assume that aside from people around the landing area, there are no other direct threats to the landing procedure.

\subsection{Adaptive emergency landing}
\label{subsec:adaptiveland}
\begin{figure}[!htb]
	\centering
	\includegraphics[width=0.8\textwidth]{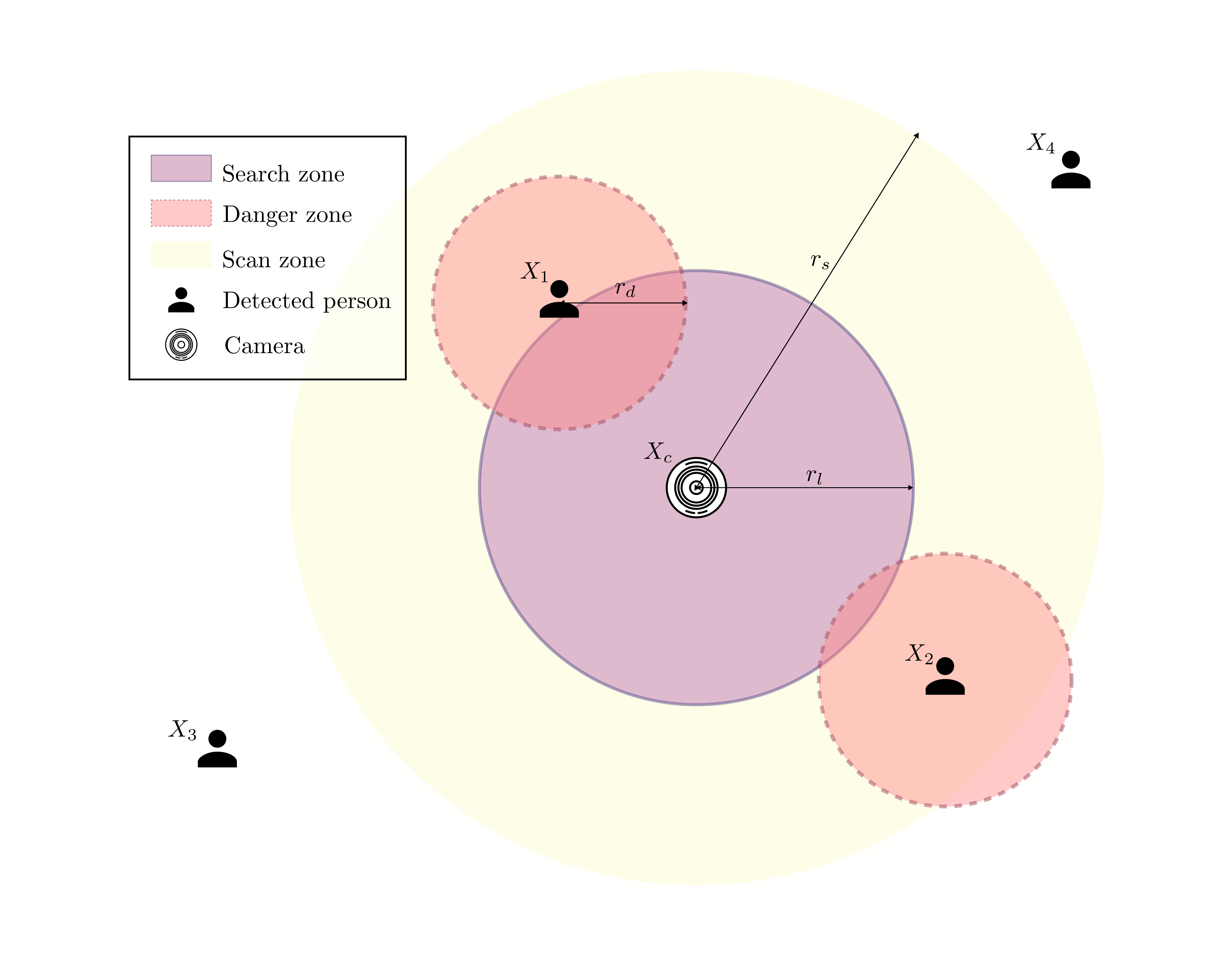}
	\caption{Optimization problem to solve to obtain safe landing position.}
	\label{fig:optimprob}
\end{figure}

When the vehicle can no longer hover at a safe position and must land immediately, the optimal position for landing, $X_o = \{x_o, y_o\}$), considering the vehicle's current position, which is also the camera's position, must satisfy multiple criteria. Firstly, it must move as far away from the surrounding people as possible. Secondly, its landing position must be away from each person by a specific range. Last but not least, we must ensure that the landing position is within a certain threshold, so the search range needs to be limited. To satisfy all requirements mentioned above, we reformulate the optimal landing position search into an optimization problem and utilize a solver to get the results. We implemented the landing spot search with SciPy's \texttt{minimize} function with the optimization method Sequential Least Squares Programming (SLSQP), which is suitable for constrained optimizations. Figure~\ref{fig:optimprob} illustrates how we approach the problem.

The positions of detected people are denoted as $X_1,..., X_{n_p}$, and the position of the camera, which is also the hovering position of the UAV, is denoted as $X_c$. We define a \textit{search zone} as a circular area with radius $r_l$ where the optimizer can search for a landing spot. Around each detected person is a \textit{danger zone} with range $r_d$, which the vehicle should avoid. Finally, the \textit{scan zone} with range $r_s$ is where all the people are considered to be in danger and should be avoided. The \textit{scan zone} is also the area in which the emergency state for the flight controller is triggered, causing it to retreat the vehicle to a safe position and hover initially. To ensure that the vehicle does not go out of the \textit{scan zone}, where the camera and the detection module do not provide enough information to conclude whether there are people, we restrict that $r_l \leq r_s$.

Because we want the UAV to land as far as possible from the people standing in close vicinity of the camera/UAV's initial landing spot, for $n_p$ humans detected in the \textit{scan zone}, the function that the optimization solver must maximize is as follows:

\begin{align}
	\operatorname*{argmax}_{x_o, y_o \in [-r_l, r_l]} {\sum_{i=1}^{n_p}{\left\lVert X_o - X_i\right\rVert}}
	\label{fun:optimmax}
\end{align}
Then, to ensure the solution is not within the \textit{danger zone}, the first constraint is formulated as:
\begin{align}
	\left\lVert X_o - X_i\right\rVert \geq r_d,\ \forall i \in \{0,...,n_p\}
\end{align}
Lastly, the selected landing position must be in the predefined \textit{search zone}:
\begin{align}
	\left\lVert X_o - X_c\right\rVert \leq r_l
\end{align}

While maximizing the function \ref{fun:optimmax} results in a landing position that is the furthest from all detected people, it is sometimes safer to emphasize the people who are closer to the camera, which is also the hovering position of the UAV. To do so, we introduce another term to address how close a person is to function \ref{fun:optimmax}, and rewrite it as:
\begin{align}
	\operatorname*{argmax}_{x_o, y_o \in [-r_l, r_l]} {\sum_{i=1}^{n_p}{\frac{1}{{\left\lVert X_i - X_c
				\right\rVert}^\alpha}\left\lVert X_o - X_i\right\rVert}}
\end{align}
The parameter $\alpha$ controls how much the distance of each detected person in the \textit{scan zone} to the UAV's current position impacts the selection of the landing spot. In other words, the higher $\alpha$ is, the more the UAV tries to avoid people close to it.

\subsection{Offboard navigation}
Algorithm~\ref{alg:safelanding} summarizes the safe landing program on the UAV's companion computer. It amalgamates the visual-based localization algorithm, the adaptive emergency landing algorithm, and a simple finite state machine determining each mission phase's setpoint. Several parameters related to the flight mission must be pre-determined, including the take-off height, the mission waypoints, and the pre-landing height above the landing pad. Furthermore, the parameters mentioned in Section~\ref{subsec:adaptiveland} should be tuned for different situations.

The condition mentioned in line~\ref{algline:cond} of the algorithm is used for the simulated and indoor experiments, which only sets a timeout period for the UAV's hovering. This condition can be extended to adapt to more types of emergencies that require immediate landing.

\begin{algorithm}
	\caption{Safe landing algorithm.}
	\label{alg:safelanding}
	\DontPrintSemicolon
	\SetKwFunction{Bboxcb}{bbox\_callback}
	\SetKwProg{Subs}{Subscribe to}{:}{}
	\SetKwProg{Publ}{Publish to}{:}{}
	\SetKwProg{Fn}{Function}{:}{}

	\Fn{\Bboxcb{msg}}{
		\For{$box \in msg.bounding\_boxes$}{
			\If{$box.dist \leq r_s$}{
				$EMERGENCY \gets True$\;
				$danger\_counter++$\;
				break\;
			}
		}
		\tcc{If the vehicle has hovered for too long, switch to adaptive emergency landing}
		\If{$danger\_counter \geq danger\_threshold$}{\label{algline:cond}
			\If{$not\ EMERGENCY\_LANDING$}{
				$X_{ppl} \gets [\ ]$\tcc*[r]{List of people in the scan zone}
				\For{$box \in msg.bounding\_boxes$}{
					\If{$box.dist \leq r_s$}{
						$X_p \gets$ localize(box)\tcc*[r]{Section~\ref{subsec:vislocal}}
						$X_{ppl}$.append$(X_p)$\;
					}
				}
				$X_o \gets$ optim\_search$(X_{ppl}, x_s, x_r, x_l,
					\alpha)$\tcc*[r]{Section~\ref{subsec:adaptiveland}}
				$prelanding\_pos\left[:2\right] += X_o$\;
				$EMERGENCY\_LANDING \gets True$\;
			}
		}
	}
	\While{$True$}{
		\Switch{$FLIGHT\_MODE$}{
			\Case{$TAKEOFF$}{
				$setpoint \gets takeoff\_position$
			}
			\Case{$MISSION$}{
				$setpoint \gets$ get\_mission\_setpoint($mission\_setpoints$)
			}
			\Case{$PRELANDING$}{
				\eIf{$EMERGENCY$ and not $EMERGENCY\_LANDING$}{
					$setpoint \gets retreat\_pos$\tcc*[r]{Hover and wait}
				}
				{
					$setpoint \gets prelanding\_pos$\tcc*[r]{Default or emergency landing}
				}
			}
			\Case{$LANDING$}{
				$setpoint \gets None$
				land\_command()
			}
		}
		\If{$setpoint$ != None}{
			publish\_setpoint($setpoint$)
		}
	}
\end{algorithm}

\section{Experimental Results}

\subsection{Data preparation}
The inputs for the detector described in Section~\ref{sec:detmod} are image frames collected from a single PICAM360 panoramic camera module. The images are retained in their original circular panoramic form to minimize the amount of pre-processing required and streamline implementation on embedded platforms. Our dataset comprises a training set with 5,062 images from 7 ROS bags and a test set with 2,030 images from 2 ROS bags. 

To further enrich the dataset, data augmentation is a viable option that has proven effective in improving deep learning models' performance in various domains, including computer vision~\cite{shorten2019augmentdatasurvey,kaur2021augmentobjdet}. We applied rotational transformations to the original training and test sets with angle $\theta \in [90\degree, 180\degree, 270\degree]$. Conventionally, augmenting the test set is not advisable because it is crucial to maintain the authenticity of the unseen data that the model might encounter in the real world. However, rotating a circular fisheye image is valid in the context of this work as it simulates changes in the camera placement angle, which is very likely to happen in our application. Unlike pinhole cameras that have to be upfront when taking regular photos, the panoramic camera in this setup does not have to be in any specific orientation. Furthermore, while people are moving around the camera when the dataset is collected, the background does not, so hypothetically, this will also enhance the robustness of the trained models.

Another requirement for this project is estimating the distance between each detected human and the camera module. We experimented with two methods to get this information: Decawave's ultra-wideband (UWB) module DWM1001 and the Optitrack MOCAP system. Each person in the experiment holds a UWB module or a set of reflective markers; another module will be placed where the camera is. The distance between the person to the camera is calculated as the Euclidean distance in two-dimensional space between the module they are holding and the camera.

\subsection{Human detector training details}
\label{sec:detectortrain}
\subsubsection{Evaluation metric}
The trained object detector should be able to reliably detect potential hazards to the landing operation, in this case, people around the landing site. Detecting people near the landing site is more critical as they pose a direct danger to the operation while simultaneously the UAV poses a hazard to nearby passers. Because the dimensions of the experimental flight zone are $9 \times 8\,m^2$, and the camera is placed near the middle of it during data collection and slightly shifted to the side during experiments, the furthest possible distance to the camera is approximately 6\,m. We select 3\,m as the safe range for the experiment, i.e., the distance that a detected person is considered close to the camera. To evaluate the trained model's performance on people at different ranges, we use a slightly modified version of COCO's AP across object size metric. Examining the distance data illustrated in Figure \ref{fig:distarearel} shows that this distance negatively correlates with the normalized area of the corresponding bounding box, so it is reasonable to use the box area as a rough estimate to separate instances that are close to or far from the camera. The median bounding box area of samples within $3\pm 0.2$\,m vicinity of the camera from the dataset is approximately $0.0135$, so the metrics that we use are: 
\begin{itemize}
	\item $\mathbf{AP^{F}}$: AP for far objects, bounding boxes with $area \geq 0.0135 h_{im}w_{im}$
	\item $\mathbf{AP^{N}}$: AP for near objects, bounding boxes with $area \leq 0.0135 h_{im}w_{im}$
	\item $\mathbf{AP^{all}}$: AP for all objects
\end{itemize}

\subsubsection{Fine-tuning on panoramic dataset}

To leverage the well-initialized weights of the pre-trained models, we use them as the foundation and fine-tune them on our training set. With this technique, it is possible to obtain a model capable of performing in the target environment with a relatively small amount of data compared to the large-scale COCO dataset. As mentioned in Section~\ref{subsec:yoloimp}, we focus on training two model versions, YOLOv7-tiny and YOLOv7. Both models are trained with base inference size 640 and multi-resolution. Multi-resolution training is a technique that varies the training resolution to $\pm50\%$  of the base resolution, which should improve the model's robustness to scaling changes and prediction performance on small objects. Previous research has reported promising results on this training technique's effectiveness in improving object detection models' resolution scalability~\cite{tian2022resformer,yan2013multirespedes}.

\subsubsection{Ablation studies}
\label{subsec:ablationobjdet}

We now analyze the effect of data augmentation and multi-resolution training in the performance of the trained model.

\textbf{Rotational augmentation}

When forming the dataset, we apply rotational augmentation to enrich the dataset. We conducted this ablation study by evaluating the performance of the fine-tuned models on both the unaugmented and augmented datasets. Table~\ref{tab:abla} shows that the models trained on the augmented dataset outperform those trained only on unrotated data in all test cases.

\begin{table}[!htb]
	\caption{Performance of the fine-tuned (FT) models and ablation study on the effect of multi-resolution (MR) training and training on rotationally augmented data (Aug).}
	\label{tab:abla}
	\centering
	\setlength{\tabcolsep}{0.001pt} 
	\begin{tabularx}{\textwidth}{Bttttttt}
		\toprule
		\textbf{Model}                   & $\mathbf{AP^{all}}$ & $\mathbf{AP^{N}}$                                 & $\mathbf{AP^{N}_{50}}$ & $\mathbf{AP^{N}_{75}}$ & $\mathbf{AP^{F}}$ & $\mathbf{AP^{F}_{50}}$ & $\mathbf{AP^{F}_{75}}$ \\
		\midrule
		                                 &                     & \multicolumn{6}{c}{\textbf{Unaugmented test set}}                                                                                                                         \\
		\cmidrule(lr){2-8}
		YOLOv7-tiny (\textbf{baseline})  & 0.068               & 0.173                                             & 0.361                  & 0.115                  & 0.019             & 0.041                  & 0.013                  \\
		YOLOv7-tiny (\textbf{FT})        & 0.316               & 0.529                                             & 0.937                  & 0.531                  & 0.238             & 0.490                  & 0.201                  \\
		\textbf{YOLOv7-tiny (FT+Aug)}    & \textbf{0.388}      & \textbf{0.528}                                    & \textbf{0.955}         & 0.527                  & \textbf{0.327}    & \textbf{0.603}         & \textbf{0.337}         \\

		\midrule[0.2pt]

		YOLOv7 (\textbf{baseline})       & 0.133               & 0.321                                             & 0.661                  & 0.263                  & 0.045             & 0.117                  & 0.032                  \\
		YOLOv7 (\textbf{FT})             & 0.393               & 0.535                                             & 0.941                  & 0.545                  & 0.327             & 0.664                  & 0.301                  \\
		\textbf{YOLOv7 (FT+Aug)}         & \textbf{0.415}      & \textbf{0.572}                                    & \textbf{0.961}         & \textbf{0.615}         & \textbf{0.343}    & \textbf{0.601}         & \textbf{0.377}         \\
		\midrule
		                                 &                     & \multicolumn{6}{c}{\textbf{Augmented test set}}                                                                                                                           \\
		\cmidrule(lr){2-8}
		YOLOv7-tiny (\textbf{baseline})  & 0.098               & 0.174                                             & 0.392                  & 0.118                  & 0.066             & 0.160                  & 0.044                  \\
		YOLOv7-tiny (\textbf{FT})        & 0.306               & 0.490                                             & 0.927                  & 0.453                  & 0.230             & 0.478                  & 0.182                  \\
		YOLOv7-tiny (\textbf{FT+Aug})    & 0.380               & 0.513                                             & 0.944                  & 0.519                  & 0.318             & 0.576                  & 0.340                  \\
		\textbf{YOLOv7-tiny (FT+MR+Aug)} & \textbf{0.394}      & \textbf{0.532}                                    & \textbf{0.966}         & \textbf{0.534}         & \textbf{0.334}    & \textbf{0.602}         & \textbf{0.354}         \\

		\midrule[0.2pt]
		YOLOv7 (\textbf{baseline})       & 0.187               & 0.324                                             & 0.663                  & 0.267                  & 0.126             & 0.287                  & 0.085                  \\
		YOLOv7 (\textbf{FT})             & 0.368               & 0.510                                             & 0.936                  & 0.485                  & 0.303             & 0.594                  & 0.287                  \\
		YOLOv7 (\textbf{FT+Aug})         & 0.419               & 0.560                                             & 0.974                  & 0.594                  & 0.354             & 0.630                  & 0.376                  \\
		\textbf{YOLOv7 (FT+MR+Aug)}      & \textbf{0.426}      & \textbf{0.572}                                    & \textbf{0.969}         & \textbf{0.612}         & \textbf{0.359}    & \textbf{0.629}         & \textbf{0.390}         \\
		\bottomrule
	\end{tabularx}
\end{table}

\textbf{Multi-resolution training}

We utilized multi-resolution training to improve inference accuracy and robustness to scaling changes during the training process. As shown in Table~\ref{tab:abla} this training method improves the model's performance compared to the model trained with fixed resolution.

\subsubsection{Deployment on embedded platform}
\label{subsec:deploy}
When the models are well-trained on the custom dataset and ready for deployment, they are converted to TensorRT engines and deployed on the target embedded platform, the NVIDIA Jetson Xavier NX. When integrating with the ROS detection node running on the ground computing platform, the ROS bounding box and distance messages must be published at a high frequency to address the tight latency requirements of real-time applications. The maximum frequency this message can be published is 30Hz, the highest supported framerate of the PICAM360 module. Because the XGBoost prediction has little to no effect on the topic's frequency during our experiments, the essential factor in the detection module's speed is the object detector's throughput. Table~\ref{tab:bboxpubrate} shows the frequency of the bounding box and distance messages when the ROS node is running on the target embedded computer. We gather these measurements using the \texttt{rostopic} tool.

\begin{table}[!htb]
	\renewcommand*{\arraystretch}{1.6}
	\caption{Frequency of \texttt{/yolov7/boundindboxes\_dist} ROS topic at inference time with Pytorch and
		TensorRT implementations on an NVIDIA Jetson Xavier NX}
	\centering
	\begin{tabularx}{\textwidth}{Ryyyy}
		\toprule
		\multirow{2}{\hsize}{\textbf{Object detection model}} & \multicolumn{2}{c}{\textbf{Pytorch}} &
		\multicolumn{2}{c}{\textbf{TensorRT}}
		\\
		\cmidrule(lr){2-3} \cmidrule(lr){4-5}
		                                                      & \textbf{Visual data sent}            &
		\textbf{No visual data sent}                          & \textbf{Visual data sent}            &
		\textbf{No visual data sent}                                                                         \\
		\midrule
		YOLOv7-tiny                                           & 15 Hz                                & 20 Hz
		                                                      & 28 Hz                                & 30 Hz
		\\
		YOLOv7                                                & $\sim$                               & 5 Hz
		                                                      & 18 Hz                                & 20 Hz
		\\
		\bottomrule
	\end{tabularx}
	\label{tab:bboxpubrate}
\end{table}
\subsection{Distance estimator training details}
We divided the training data into five sets of bounding box data from 5 different ROS bags. To optimize and validate the performance of the distance estimator, we keep one holdout set and perform a randomized search with cross-validation (Scikit-learn's \texttt{RandomizedSearchCV}) using the training set to obtain the optimal hyperparameters as follows:
\begin{itemize}
	\item \texttt{max\_depth}: 3
	\item \texttt{learning\_rate}: 0.05
	\item \texttt{n\_estimators}: 500
	\item \texttt{colsample\_bytree}: 0.5
	\item \texttt{colsample\_bylevel}: 0.8
	\item \texttt{subsample}: 0.6
\end{itemize}

These hyperparameters slightly improve the performance on the holdout set over the default ones, and the results in mean absolute error (MAE), median absolute error (MedAE), maximum error (MaxErr), and explained variation (ExpVar) are shown in Table~\ref{tab:trainedxgb}

\begin{table}[!htb]
	\caption{Performance comparison between XGBoost models with default and tuned hyperparameters.}
	\label{tab:trainedxgb}
	\begin{tabularx}{\textwidth}{Xyyyy}
		\toprule
		\textbf{Hyperparameters} & $\mathbf{MAE}$ & $\mathbf{MedAE}$ & $\mathbf{MaxErr}$ & $\mathbf{ExpVar}$ \\
		\midrule
		Default                  & 0.208          & 0.183            & 0.933             & 0.959             \\
		Tuned                    & 0.199          & 0.159            & 0.825             & 0.961             \\
		\bottomrule
	\end{tabularx}
\end{table}

\subsection{Evaluation of vision-based localization algorithm}
\begin{figure}[!htb]
	\centering
  \includegraphics[width=.99\textwidth]{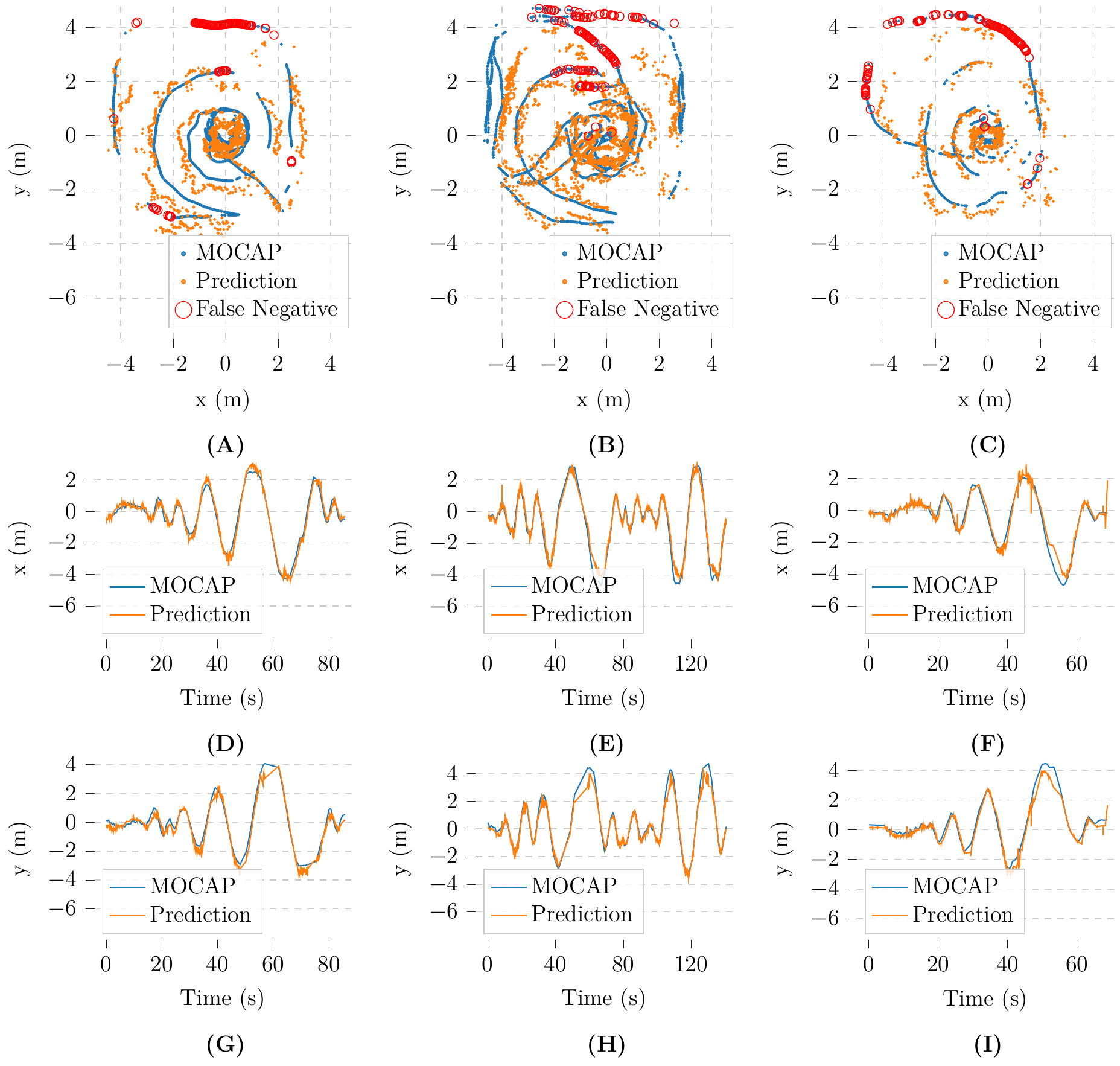}
	\caption{Vision-based localization trajectories in comparison with the trajectories from the Optitrack system (missed detections are omitted).}
	\label{fig:vislocal}
\end{figure}

As both components of the detection module have been trained and tested, we proceed to evaluate the performance of the localization algorithm base on visual information. To simplify the evaluation process, we select three datasets with one person walking around the camera for testing. Three temporary XGBoost models were also trained without the test sets to avoid high accuracy due to overfitting. The object detection model used for this test was the YOLOv7-tiny. After applying the algorithm mentioned in Section~\ref{subsec:vislocal}, the resulting trajectories are recorded and visualized in Figure~\ref{fig:vislocal}.

\begin{figure}[!htb]
	\centering
	\begin{tabularx}{0.65\textwidth}{XYY}
		\toprule
		\textbf{Trajectory} & $\mathbf{APE}$ & $\mathbf{Cossim}$ \\
		\midrule
		$\mathcal{R}1$      & 0.369          & 0.922             \\
		$\mathcal{R}2$      & 0.337          & 0.963             \\
		$\mathcal{R}3$      & 0.313          & 0.941             \\
		\bottomrule
	\end{tabularx}
  \includegraphics[width=0.65\textwidth]{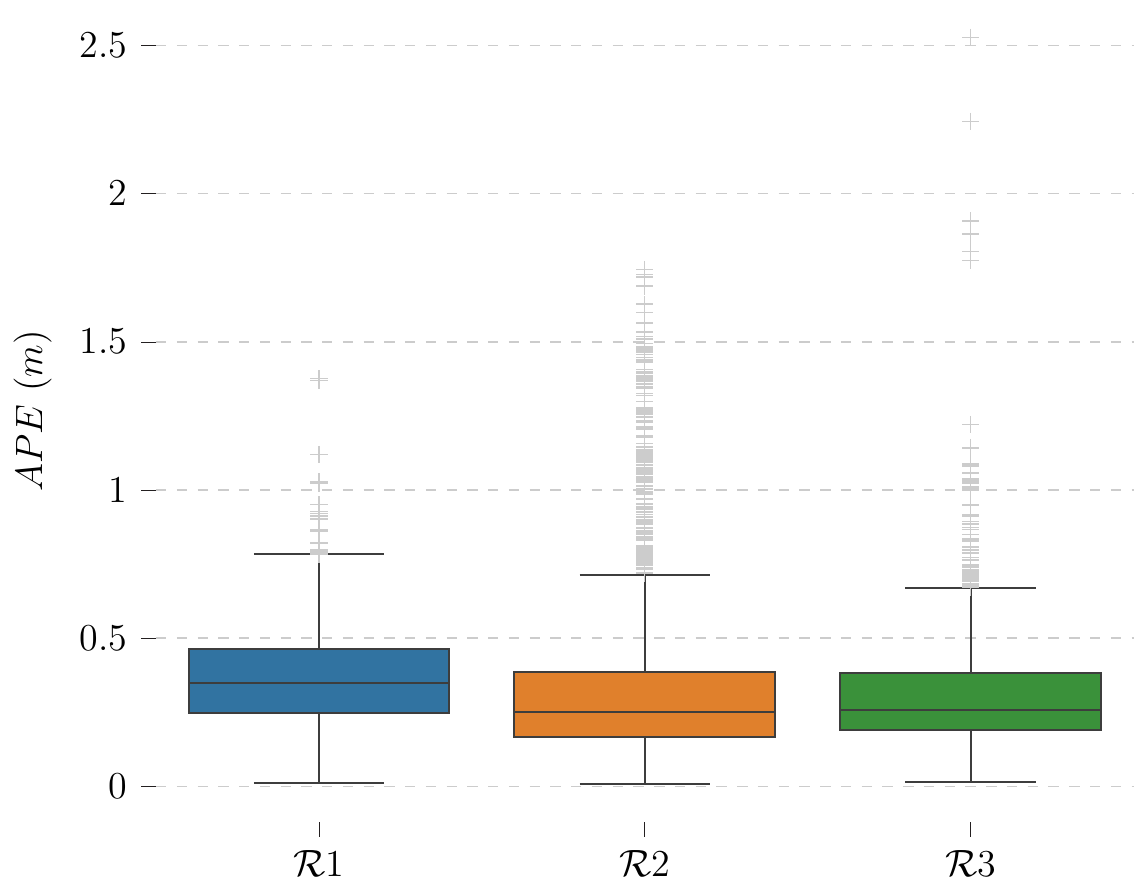}
	\caption{Evaluation of vision-based localization method. Each of the labels $\mathcal{R}_i$ represent three different experiments.}
	\label{fig:vislocperf}
\end{figure}

We evaluate these results with cosine similarity (Cossim) and average positioning error (APE) metrics. The former gives insight into how accurate our method is at determining the direction in which the person is with respect to the camera. The latter quantifies how accurate the predicted trajectories in Figure~\ref{fig:vislocal} are. The experimental results are shown in Figure~\ref{fig:vislocperf}. For simplicity, the missing bounding boxes and the frames without corresponding Optitrack data are omitted when calculating the metrics.

\subsection{Autonomous landing experiments}
\subsubsection{Simulation}
\label{subsec:simulation}
The autonomous flight programs are thoroughly tested in a simulation environment before deployment to guarantee safety. The simulation environment was implemented with PX4 Gazebo SITL. The tests are conducted on a Laptop with an NVIDIA RTX3070 GPU to run the YOLOv7 models on image frames from the PICAM360. From the simulated results, we validate that both designed behaviors, hovering and adaptive emergency landing, function correctly during the pre-landing phase.

\subsubsection{Experiments with real UAV}
\label{subsec:demo}
\begin{figure}[!htb]
	\centering
	\includegraphics[width=0.65\textwidth]{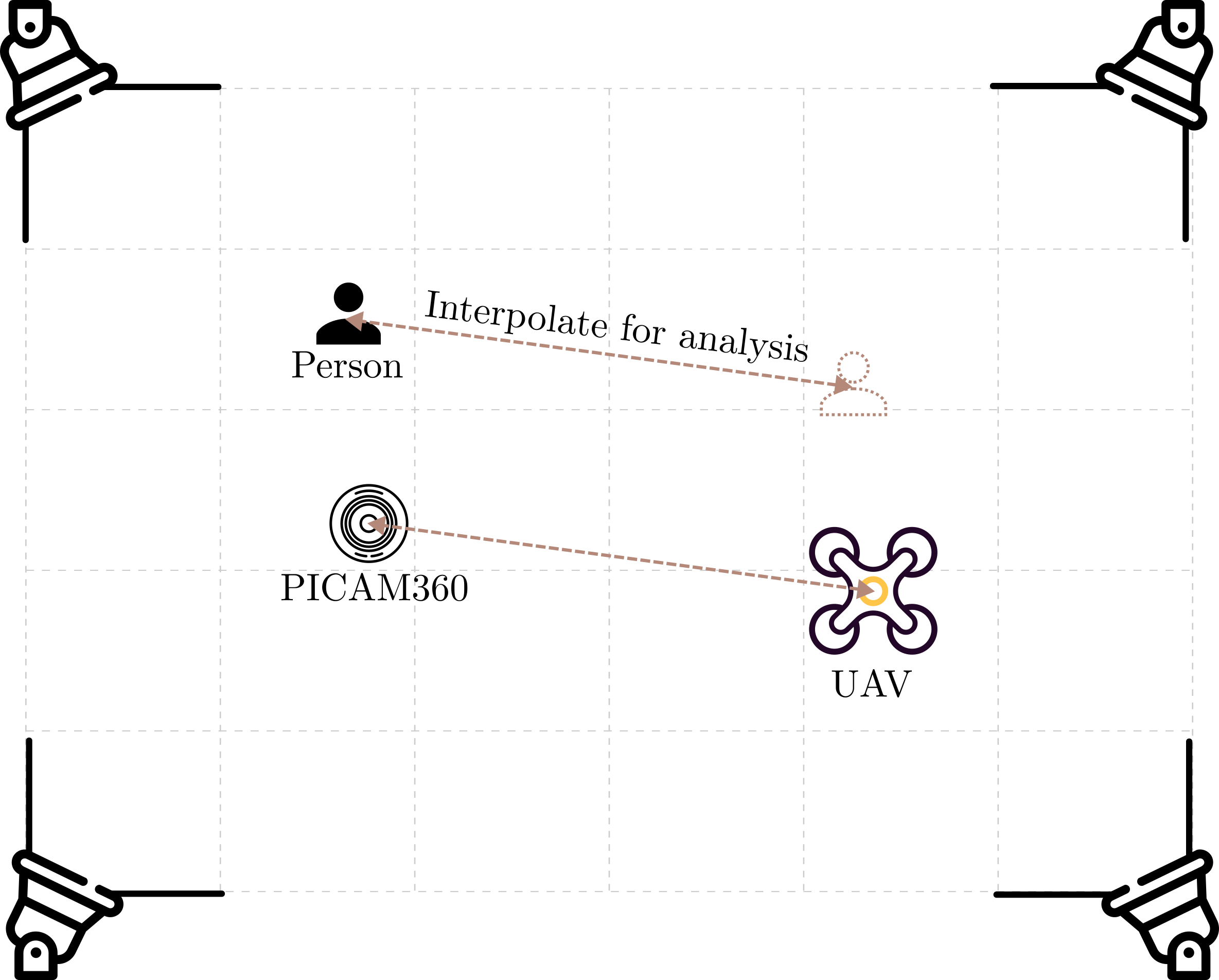}
	\caption{Experimental setup. For safety reasons during the experiments, the camera is not located directly where the drone is landing. This allows for more experimental setups while staying at a safe distance from the UAV.}
	\label{fig:expsetup}
\end{figure}

We conduct experiments within our flight zone to test the system's performance. The object detection model used in the detection module is YOLOv7-tiny. The parameters for the emergency landing algorithm are: $r_l=1m,\ r_s=3m,\ r_d=0.5m$, and $\alpha=0$ (see Section~\ref{subsec:adaptiveland} for more information on the algorithm). We deliberately choose a small range for the \textit{search zone} to keep the experimental UAV within the flight zone. Furthermore, to simplify the experiment, the mission only consists of the UAV taking off and landing at the same spot afterward because we are most interested in the latter's behavior for the scope of this thesis. Because the safe landing software is still in development, to ensure the safety of the people involved in the experiments, as well as to protect the equipment of the on-ground monitoring system, we place the embedded computer running the detection module and the panoramic camera away from the drone during the experiment and interpolate the positions of the camera and people to the drone's position while analyzing the results. The experimental setup is described in Figure~\ref{fig:expsetup}

\begin{figure}[!htb]
	\centering
	\includegraphics[width=\textwidth]{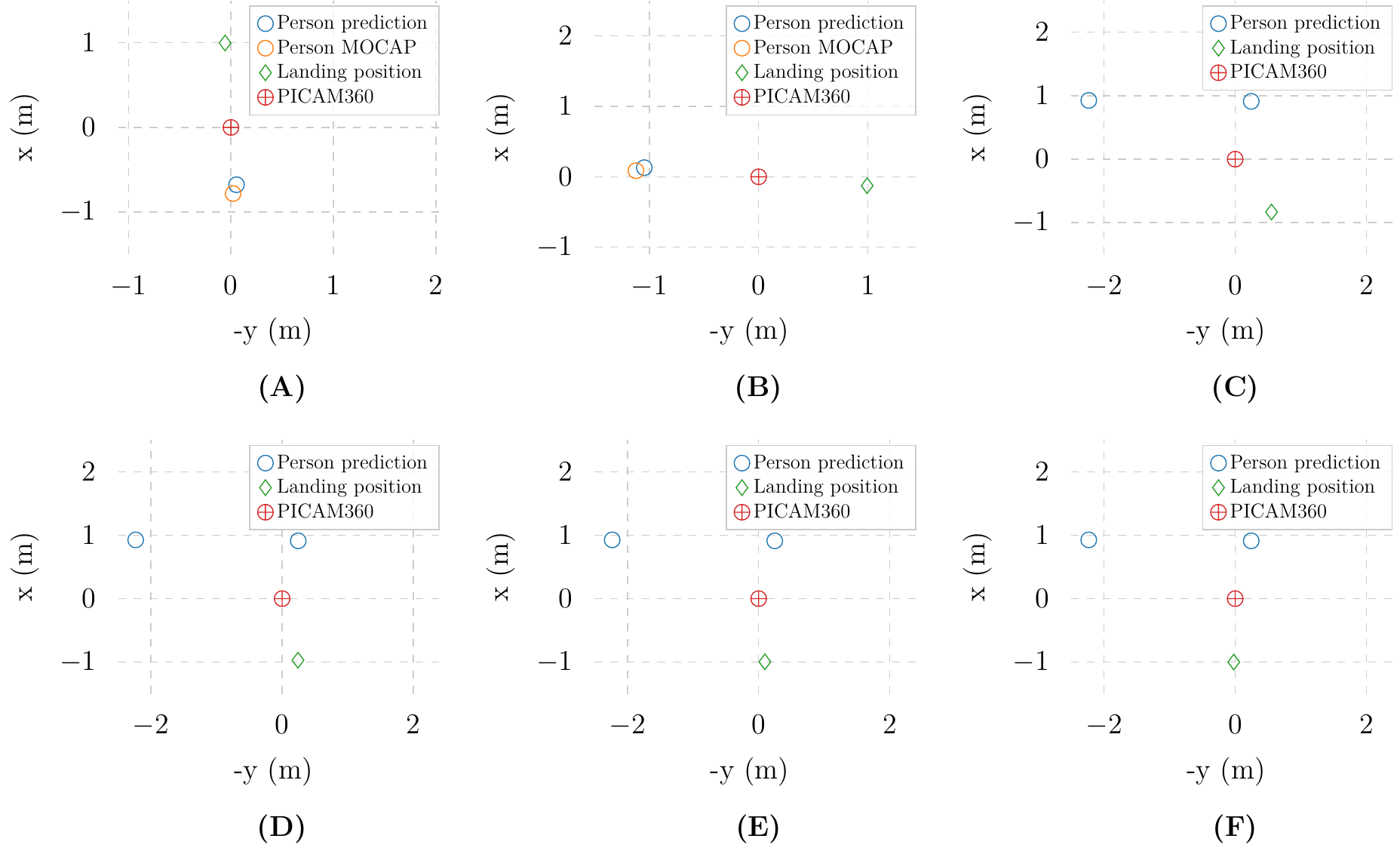}
  \caption{Optimal landing positions in different experiments with (A-B) one person approaching and (C-F) two people aproaching: (C) $\alpha=0$ (D) $\alpha=1$ (E) $\alpha=1.5$ (F) $\alpha=2$}
	\label{fig:safeland}
\end{figure}

We experimented with one and two people approaching the camera in multiple directions. The hovering and adaptive emergency landing behaviors work as expected in all experiments. Figure~\ref{fig:safeland} presents the optimal landing position selection results for clearer visualization. In Figure~\ref{fig:safeland}C, while the optimal landing position maximizes the distance to the detected people, it does not prioritize the closest person to the drone. As explained in Section~\ref{subsec:adaptiveland}, this behavior can be altered by modifying the parameter $\alpha$ in Function~\ref{fun:optimmax} to increase the prioritization of the algorithm on the distance of the detected person to the camera/drone's position. This effect is demonstrated in Figure~\ref{fig:safeland}C-F, which shows that increasing $\alpha$ increases the distance between the landing position and the closest person.

The experiments, both in simulation and in real-world indoor environments, are recorded in a video and uploaded at \url{https://www.youtube.com/watch?v=XdolUS1bUVs}

\section{Conclusion}

In this paper, we propose a novel on-ground vision-based solution for safe UAV landing by leveraging the omnidirectional view capability of panoramic sensors. The detection module, comprising a YOLOv7-based object detector and an XGBoost-based distance estimator, demonstrates high capability in detecting and localizing humans near the landing zone while delivering real-time performance. Furthermore, a series of indoors experiments has proven the system's reliability in enabling landing UAVs to avoid surrounding pedestrians. Rather than completely replacing available onboard methods~\cite{marcu2018safeuav, tovanche2022visual}, our solution serves as an extra layer of safety for UAV landing applications. Our ultimate goal is a collaborative autonomy approach where sensor and detection data from the micro-airports is fused with the UAVs' sensors and computational capabilities to enhance the system's reliability, safety, and efficiency.


\nocite{*}
\bibliographystyle{unsrt}
\bibliography{bibliography}

\end{document}